\documentclass{article} % For LaTeX2e
\usepackage{iclr2026_conference,times}

\usepackage{amsmath,amsfonts,bm}

\def\eqref#1{equation~\ref{#1}}
\def\1{\bm{1}}

\DeclareMathAlphabet{\mathsfit}{\encodingdefault}{\sfdefault}{m}{sl}
\SetMathAlphabet{\mathsfit}{bold}{\encodingdefault}{\sfdefault}{bx}{n}

\usepackage{hyperref}
\usepackage{url}
\usepackage{graphicx}
\usepackage{subcaption}
\usepackage{wrapfig}

\title{Representing expertise accelerates learning from pedagogical interaction data}
\author{Dhara Yu, Karthikeya Kaushik, Bill D. Thompson  \\
UC Berkeley \\
\texttt{\{dharakyu, karthikeya.kaushik, wdt\}@berkeley.edu} \\
}

\iclrfinalcopy % Uncomment for camera-ready version, but NOT for submission.
\begin{document}

\maketitle

\begin{abstract}
%Recent work has suggested that training LLMs on traces of interactions between other agents can improve performance, yet it remains unknown which features of the interactions contribute to these improvements. 
Work in cognitive science and artificial intelligence has suggested that exposing learning agents to traces of interaction between multiple individuals can improve performance in a variety of settings, yet it remains unknown which features of interactions contribute to this improvement.
We examined the factors that support the effectiveness of interaction data, using a controlled paradigm that allowed us to precisely operationalize key distinctions between interaction and an expert acting alone.
We generated synthetic datasets of simple interactions between an expert and a novice in a spatial navigation task, and then trained transformer models on those datasets, evaluating performance after exposure to different datasets. Our experiments showed that models trained on pedagogical interactions were more robust across a variety of scenarios compared to models trained only on expert demonstrations, and that having the ability to represent epistemically distinct agents led to expert-like behavior even when expert behavior was rarely observed.
%We found that training on traces of \textit{pedagogical} interaction, in which an expert corrects suboptimal behavior of a novice, leads to improved performance in situations that an expert acting alone would be unlikely to encounter, and that training on interaction confers additional benefits when the training data reveals which agent generated a sequence of behaviors. These results highlight structural properties of interactions that lead to improved robustness. 
\end{abstract}

\section{Introduction}

Much of what we know about the world comes from observing interactions between other people \citep{chuey2025theory}. Examples of this form of learning range from listening to a question-and-answer session between a speaker and an audience member, to reading a comment thread on a cooking blog. Even young children are capable of learning from third-party observations, showing the ability to infer complex causal structure from overheard conversations \citep{bonawitz2011double}. Observing interactions between others has been argued to be crucial to the acquisition of language itself, as these interactions constitute a more substantive portion of child language data than has traditionally been assumed \citep{foushee2024infants}.

There is a striking parallel between this setting in human learning and large language model pretraining on internet corpora. Analogous to a child overhearing a conversation, LLMs are exposed to traces of interaction generated by other agents (e.g. in online forums), where the learning agent itself is not a participant in the interaction. 
Exposing LLMs, both in-context and through fine-tuning, to \textit{additional} traces of interaction can improve performance in a variety of tasks, such as formal reasoning \citep{du2023improving, subramaniam2025multiagent}, reading comprehension \citep{khan2024debating}, %reference games \citep{Chen2024:respect} 
and moral dilemmas \citep{liu2023training}.

These findings raise the question of what structural properties of interaction might contribute to improved performance in LLMs.
Work in cognitive science has suggested that learning from interactions might be useful because interaction exposes uniquely informative content beyond what is surfaced by a single agent acting alone, such as corrective feedback and recovery from mistakes \citep{fox1999listening}. This problem has also been studied within the robotics community, motivating approaches such as DAgger, which captures the idea that novice exploration exposes more of the state space \citep{ross2011reduction}. In human learners, corrective events are most likely to happen in interactions with knowledge asymmetries, e.g. \textit{pedagogical} interactions between novices and experts, in which experts intervene on suboptimal behavior from novices. Cognitive models also suggest that for learning agents to benefit from these rich forms of data, it helps to be endowed with social reasoning capabilities, such as an ability to represent data-generating agents' differing levels of expertise \citep{shafto2014rational, landrum2015learning}. 

Here we evaluated if these factors also modulate the effectiveness of interaction data for language models, %that learning from these types of interactions can improve generalization,
using a controlled paradigm inspired by past work characterizing representations in LLMs \citep{vafa2024evaluating}. 
We generated synthetic datasets of agents interacting in a spatial planning task using symbolic planning algorithms, which captured the core sequential structure of natural language interaction. We manipulated the structure of the generated examples to capture the key distinctions between experts acting alone and multiple individuals interacting, focusing on a particular type of interaction: experts intervening on the behavior of a novice. We also manipulated the datasets to encourage distinct representation of agent types. Then, we trained transformer language models on these datasets and evaluated models' performance on the task in a variety of scenarios.

Our results revealed that models trained on traces of interaction were more robust in scenarios that an expert acting alone would be unlikely to encounter.
Furthermore, we found that interaction data provided an even more expansive benefit when there were clear indicators of which agent (e.g., novice or expert) generated the data.
This highlights potential mechanisms through which training on interaction data yields improved performance, and raises important questions for future work.

\section{Methods}

\begin{figure*}
    \centering
    \includegraphics[width=0.8\textwidth]{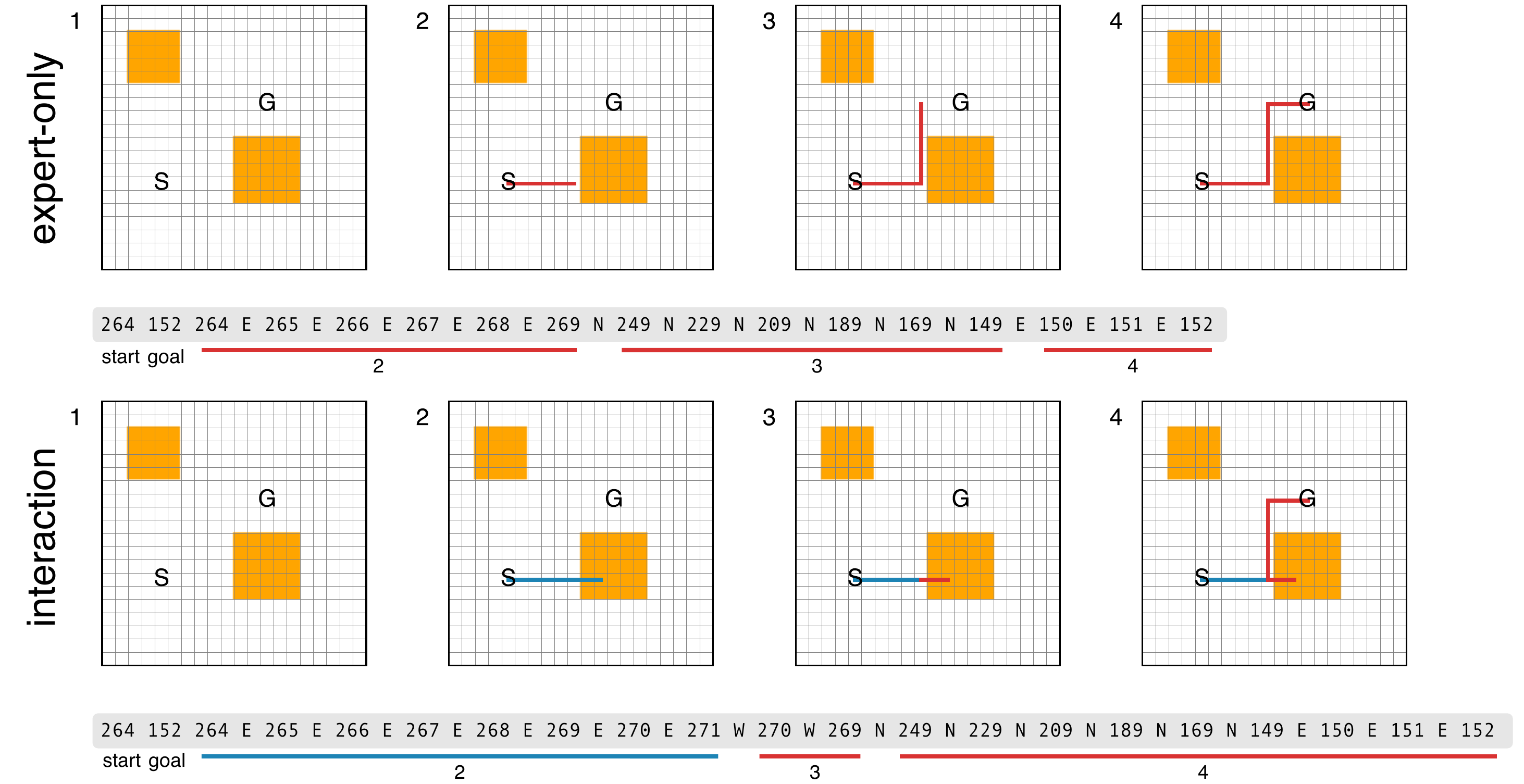}
    \caption{Example traces under expert-only and interaction policies. Grids show the agent's trajectory at 4 different time steps. Sequences are the representations of the full trajectory seen by the model, with the underlines indicating the part corresponding to the sub-trajectories at each timestep.}
    \label{fig:overview}
    \vspace{-10pt}
\end{figure*}

\subsection{Simulating interactions}
Our goal is to study when learning from interactions between novices and experts can lead to distinct benefits over learning from an expert acting alone. To do so, we needed a setting that allowed us to generate controlled datasets of \textit{expert-only} and \textit{interaction} behaviors, which could then be used as training data for a transformer. We generated this data in the context of a generic planning task. The goal of the task was to find the optimal path between a start point and a goal point, in a particular grid environment. Each grid contained certain high-cost cells; to maximize reward in the task, agents needed to take the most direct path to the goal that avoided costly cells. This task was suitable because interaction could be operationalized in a meaningful way, for example through correcting the trajectory of a novice. While simplified, this setting captures the core dynamics of pedagogical interactions unfolding over time in natural language datasets, e.g. Stack Overflow threads in which expert programmers help novices debug.

Here, we describe the procedure for generating datasets.
We formulated the task as an MDP, where the set of states is defined by a $20 \times 20$ grid, with set of high-cost states $H$ and a goal state $g$. The reward function $R_{H, g}$ assigns $+100$ for reaching $g$, $-20$ for states in $H$, and $-1$ otherwise (see Appendix~\ref{app:mdp} for formal definitions).
A trajectory $\tau$ is an alternating sequence of states and actions representing a path from the start state $s_0$ to the goal state $g = s_T$: $\tau = (s_0, a_0, s_1, a_1, \dots, a_{T-1}, s_T)$.

Next, we describe the expert-only and interaction policies used to generate trajectories:
%\subsubsection{Expert-only policy}
%For a given start/goal pair in a given grid, we used the value iteration algorithm to find the optimal path (for each unique goal location, we modified the reward function such that reaching the goal incurred a reward of +100). The resulting path was represented as a series of tokens, comprised of the start location, the goal location, followed by the complete trajectory of movements between the start and the goal (see figure). Each unique cell, as well as each cardinal direction, had a token in the vocabulary of the language model.
The \textit{expert-only} policy $\pi^{*}$ represents the optimal policy under the true reward function $R_{H, g}$ (Figure \ref{fig:overview}, top). %Concretely, $\pi^*(a \vert s; H, g)$ gives the probability of taking action $a \in \mathcal{A}$, given high-cost states $H$ and goal state $g$. We used value iteration to solve for the optimal policies.
%\subsubsection{Interaction policy}
The \textit{interaction} policy $\pi_{\text{int}}$ represents the behavior of a novice acting under an incorrect reward function, who is then corrected by an expert if behavior diverges from what would be optimal under the true reward function. Intuitively, interaction is operationalized as the expert modifying the policy of the novice in a way that makes their ensuing behavior expert-like. 
%This could approximate a range of pedagogical situations, from high-fidelity transmission of a policy (e.g., describing an exact sequence of moves to make when rock climbing, and the amateur following that sequence), to instances of an expert overriding the actions of a novice (e.g, a senior doctor taking over a procedure after an error by a medical resident).
First, we define the initial novice policy $\pi_{N}$ as the optimal policy under incorrect reward function $R_{\text{naive}}$ which does faithfully capture the intended goal state $g$ but ignores the set of high-cost states $H$. %The probability of taking action $a$ in state $s$ is therefore given by $\pi_N(a \vert s; g)$. Intuitively, an agent acting under this policy takes the geometrically shortest path, regardless of high-cost states. 
The interaction policy $\pi_{\text{int}}$ switches from $\pi_N$ to $\pi^*$ at timestep $t^*$, the timestep at which the agent has been in a high-cost state for 2 consecutive actions (Figure \ref{fig:overview}, bottom). If this condition is not met (because the trajectory under the novice policy incidentally avoids high-cost states), then no intervention occurs. %Thus the interaction policy $\pi_{\text{int}}$ can be expressed as follows:
%\begin{equation*}
%    \pi_{\text{int}}(a \mid s_t; H, g) = 
%        \begin{cases} 
%        \pi^*(a \mid s_t; H, g) & \text{if } t \geq t^* \\
%        \pi_N(a \mid s_t; g)              & \text{otherwise}
%        \end{cases}
%\end{equation*}

 %\dy{need some help here! haven't really seen much cogsci work, except for Mark's paper (which does not seem widely known) that operationalizes teaching interactions in this way. there are some algorithms in the DAgger family that kind of do this, e.g. the expert (human overseer) corrects the novice (the learning agent) by just overriding the policy at some point.}

%\subsubsection{Constructing traces}
To construct training datasets, we generated a set of traces under a specific MDP (there were 10), following policy $\pi^*$ or policy $\pi_{\text{int}}$. To construct a single trace, we sampled a start location and a goal location, generated the trajectory under the designated policy, and prepended the start and goal locations to the trajectory: $(s_0, g, \tau)$ (Figure 1). We sampled start and goal locations so that neither the start nor the goal state could be a high-cost state for the given MDP. This was to capture the intuition that an expert agent would not be tasked with navigating from a high-cost state, because it would have full knowledge of the environment and consequently avoid high-cost states altogether.

\subsection{Training and evaluation}
We trained autoregressive transformer models on the traces generated under the expert-only policy and the interaction policy (see \ref{app:training_details} for additional details). To elicit model predictions, we provided the start location, the goal location and the first location of the path (which was the start location).  %Accordingly, the inclusion of the start location for the second time was necessary to elicit judgments for low-probability start locations. 

Unlike the planning algorithms used to generate training data, the learning models were not explicitly provided with the structure of the environment. Performing well on a held-out test set therefore required learning both the structure of the environment and the expert. To evaluate performance on the task, we used \textbf{exact match}, which was true of a model output if the produced trajectory was identical to the one under the ground-truth expert policy. We also used a more relaxed \textbf{correct path} metric, which was true if the trajectory was valid (e.g. it respected the allowed transitions between cells), and the trajectory ended in the goal state. This metric was diagnostic of whether a model had learned the constraints of the environment, not necessarily whether it had learned the optimal policy.

\section{Results}

\begin{wrapfigure}{r}{0.5\textwidth} % {position}{width}
    \centering
    \vspace{-12pt}
    \includegraphics[width=\linewidth]{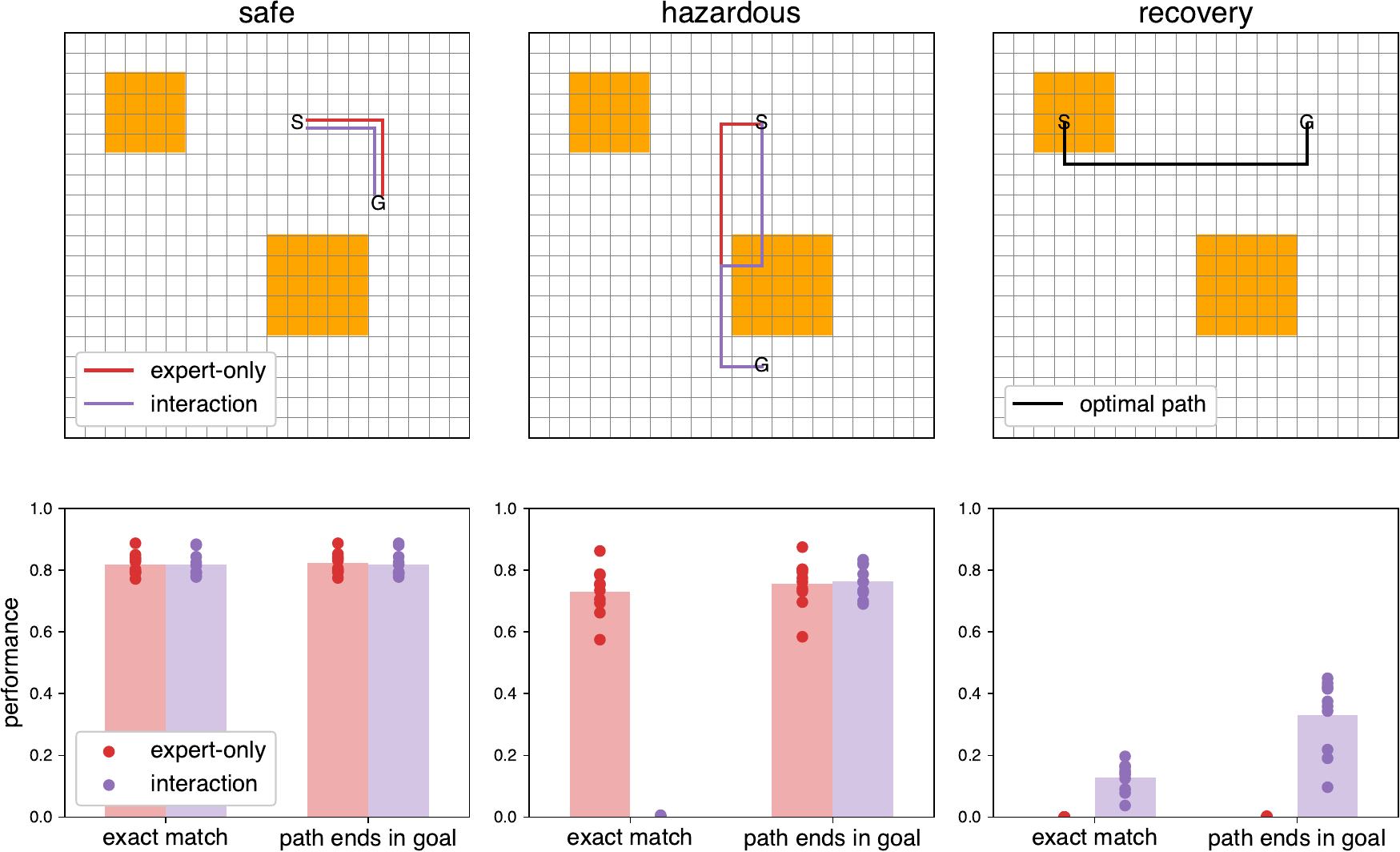}
    \vspace{-15pt} % Optional: Remove extra whitespace below caption
    \caption{Study 1 results.}
    \label{fig:experiment1}
    \vspace{-12pt} % Optional: Remove whitespace at bottom of wrap
\end{wrapfigure}

\subsection{Study 1}
To evaluate how training on different datasets affected a model's ability to produce an optimal trajectory, we constructed 3 different test sets consisting of (start, goal) pairs. In \textit{safe} trials, the provided start and goal states were not from the set of high-cost states, and the expert policy and the interaction policy both prescribed the same trajectory, meaning that the associated novice trajectories incidentally avoided entering any high-cost state. In \textit{hazardous} trials, the start and goal states were similarly not high-cost states, but following the novice policy would involve traversing through high-cost states. Finally, in \textit{recovery} trials, the start states were high-cost states while the goal states were not, probing the ability of the model to produce valid trajectories from states that would not be traversed by an expert. Each test set contained 320 held-out examples.

Figure \ref{fig:experiment1} shows model performance across the 3 test sets. In the safe trials, expert-only trained models and interaction-trained models generally performed well, confirming that in favorable conditions, models could succeed in the task. 
In hazardous trials, expert-only models generally learned to avoid high-cost states, but interaction models did not, though they did produce valid trajectories: the raw input stream of actions and states generated under an interaction was insufficient to learn the generic behavior of avoiding high-cost states. 
%The interaction-trained models generated valid trajectories, but these trajectories traversed high-cost states, and then backtracked before progressing to the goal, consistent with their training. The expert-only models, on the other hand, generally produced optimal trajectories as expected. 

Recovery trials showcased the value of learning from exploration and recovery events in interaction data. In these trials, expert-only models failed to produce valid sequences. The intervention models, on the other hand, generated optimal trajectories for some (start, goal) pairs, and for a larger percentage of coordinate pairs, produced valid trajectories. This occurred even though these models were not trained on direct examples of optimal paths between a high-cost state and a goal state.
One interpretation of the performance gap across model types is that expert-only data was not sufficiently constraining to infer that the optimal behavior in a high-cost state is to exit the state.

Overall, this pattern of results illustrates a tradeoff that arises when learning from two structurally different forms of input. Learning solely from traces of experts enables more expert-like behavior for parts of the state space that are well-traversed by the expert during learning. In contrast, learning from interaction can in principle enable recovery from suboptimal areas of the state space that an expert would not enter. %This capacity for recovery could be particularly useful in settings where the learning agent has an incomplete understanding of the domain, e.g. early on in learning, when the agent is more likely to enter suboptimal states in the first place. 
This capacity for recovery could be particularly advantageous in settings where one may need to intervene on \textit{another} agent's incorrect model, for example in teaching contexts.

%Interaction creates scenarios where you get the exploration from the novice, and the correct traces from the expert

The poor performance of interaction-trained models in hazardous trials suggests that learning from action sequences in the context of an interaction may be insufficient to acquire expert-like performance in some settings. In the next section, we explore how this can be ameliorated by training on datasets that reveal information about the agent that generated a particular sequence of actions.

\subsection{Study 2}

\begin{wrapfigure}{r}{0.6\textwidth}
    \centering
    \vspace{-12pt}
    \includegraphics[width=\linewidth]{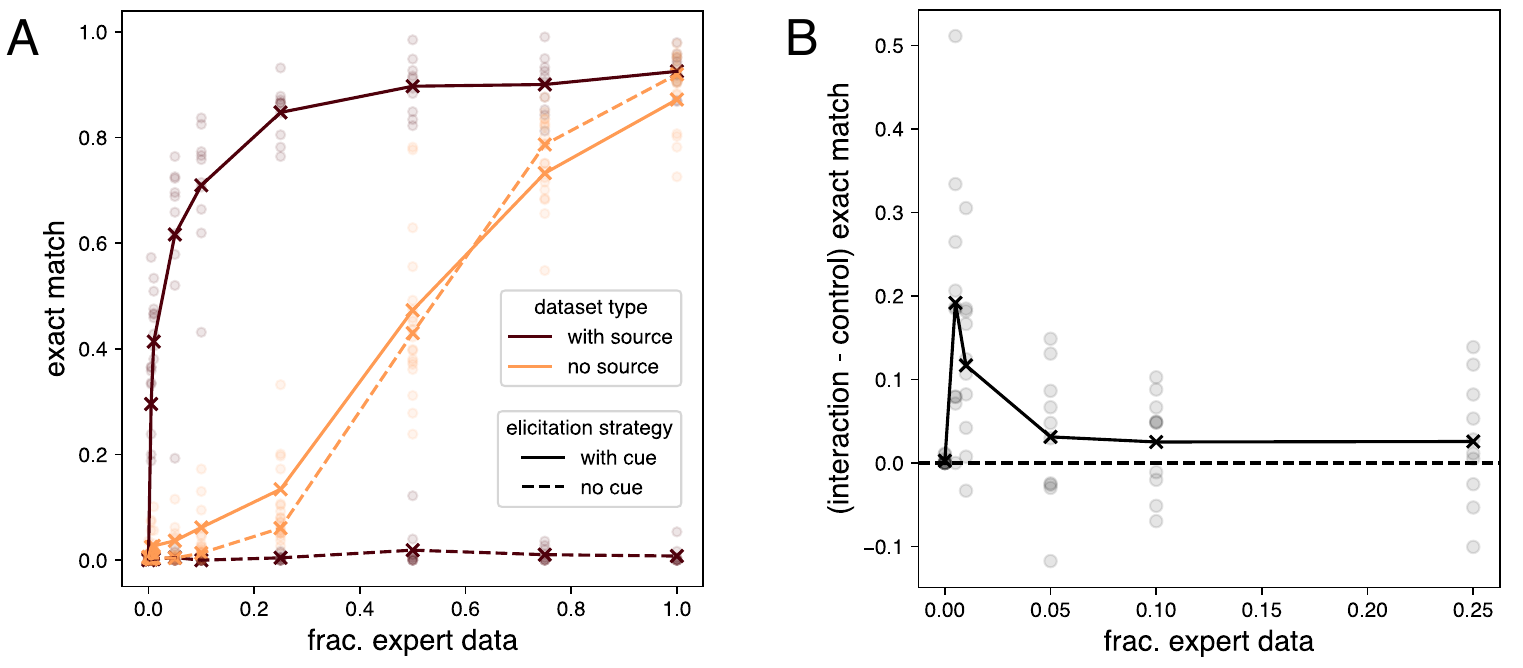}
    \caption{Study 2 results. A: performance on hazardous trials, with and without agent source indicators. B: difference in exact match accuracy (percentage points) between the models trained on interaction datasets and models trained on the matched control datasets, also on hazardous trials.}
    \label{fig:three_panels}
    \vspace{-10pt} % Optional: Remove whitespace at bottom of wrap
\end{wrapfigure}

Here our goal was to evaluate if information about the agent generating the data can improve performance in hazardous trials, through learning the generic policy of avoiding high-cost states.
To do so, we manipulated the training data to include explicit indicators of the source of certain actions in the trajectory, i.e. whether actions were brought about as a result of a novice or an expert.
Specifically, we designated source indicator tokens to prepend at the beginning of agent trajectories. An ``expert" indicator was prepended to the beginning of expert-only sequences, while a ``novice" indicator was prepended to the beginning of intervention sequences. Then, for the intervention traces in which the expert took corrective action (because the novice proceeded through high-cost states), the beginning of the corrected sequence was prepended with the ``expert" indicator (Figure \ref{fig:tagging_overview}). The tokens ``novice" and ``expert" do not have inherent semantics analogous to the meaning of those words in English; they are just indicators of 2 distinct types. This approximates scenarios in which there are outwardly visible cues in the data that indicate expertise or lack thereof. 

We hypothesized that revealing source information would be most helpful in settings where expert-only data is particularly scarce - a common situation for many specialized tasks.
The intuition is that when there are only a few samples exhibiting optimal behavior, it helps to have an explicit signal indicating which traces illustrate that behavior.
To test our hypothesis that separately representing agent types enables better generalization when exposed to limited expert-only data, we synthesized several datasets which varied in two respects (Figure \ref{fig:tagging_overview}). First, the data varied in whether the traces contained source indicators revealing the type of agent generating the trajectory (\textit{with-source}) or not (\textit{no-source}). Second, the data varied in the fraction of traces displaying expert-only or interaction behavior, ranging from containing only expert-only traces, to no traces generated under the expert policy (and being comprised solely of traces generated under the interaction policy). 

For each unique composition, we constructed a test set of hazardous trials comprised of (start, goal) pairs that had not previously been seen together in training. %We further pruned the test set such that for all examples, the exact sequence of cell coordinates and cardinal directions that comprised the optimal path for a given start/goal pair was never seen during training; this ensured that performance could not be attributed to exposure during training.
We used 2 different strategies to elicit model predictions. Under the \textit{no-cue} strategy, the model was provided with the start location and goal location, same as in Simulation 1. Under the \textit{with-cue} strategy, the model was provided with the start location, the goal location, the expert token, and then the start location again; this was done to explicitly condition the generation on an agent type (in this case, the expert type).

%The main goal of the 2x2 thing is to make it less awkward that we're using different prompting strategies for the different datasets. The most charitable framing (assuming this is borne out) is that you get a boost from the agent cue when that's in the data (that is obvious though so we shouldn't fixate on that). Then we can say there are qualitatively different patterns in the "ceiling" treatment for each of the datasets

\subsubsection{Study 2A: Learning from differentiated agents improves performance when expert-only behavior is rare}
\iffalse
\begin{itemize}
    \item The ceiling of the tagged model (e.g., taking the more favorable of the two prompting techniques) is higher than the ceiling of the untagged model. Even when expert-only traces constitute just 0.5\% of the training data, the model still gets 0.3 accuracy.
    \item The performance gap is most pronounced when the proportion of expert data is low
    \item There is no difference in the untagged model's performance between the 2 prompting techniques. In both cases, the percent of correct traces is proportional to the fraction of the data that is expert only. That is the performance ceiling, because you can't know whether you should be producing expert or novice data
    \item In contrast, there is a big gap for the tagged model. The model can make use of the expert cue in the prompt because it has seen the expert token followed by action sequences generated under a consistent expert policy. However, it fails when not provided with the scaffolding of a cue
\end{itemize}
\fi
Models trained on with-source datasets outperformed models trained on no-source datasets, under the most favorable elicitation strategies for each dataset type (Figure \ref{fig:three_panels}A).
 This performance gap was most pronounced when the proportion of expert-only data was low; even when expert-only traces constituted just 0.5\% of the training data (500 traces), models trained on with-source datasets produced optimal trajectories a mean of 30\% of the time, whereas models trained on no-source datasets virtually never produced the correct trajectory.
 %Overall, this suggests that having a basic capacity to represent distinct types leads to more sample-efficient learning.
There was no difference in the no-source models' performances with or without the expert cue. In both cases, the percent of correct traces is linearly proportional to the fraction of the data that is expert-only. This highlights a fundamental limitation of learning from no-source data in this setting: the likelihood of producing the desired behavior is bounded by the frequency of that behavior in the training data.

Models trained on with-source datasets achieve better-than-linear performance (relative to the fraction of expert-only demonstration observed during training) when provided with an expert cue, but fail entirely without this cue. This suggests that models have learned to rely on the source indicator to predict the ensuing behavior and therefore in its absence cannot do so. 
To overcome this brittleness, we ran an additional experiment in which we varied the frequency with which the source indicators appeared in the data (see \ref{app:noisy_differentiation} for details). We found that reducing the frequency of source indicators led to models reaching ceiling performance in the no-cue condition (e.g. performance was linearly proportional to the fraction of the data that was generated by an expert; Figure \ref{fig:drop_tags}). This suggests that training on partial, rather than full, source information improves robustness. 
%Overall, this pattern of results suggests that learning from data that includes the sources of actions can help learning agents make use of complex interaction data, but this benefit can be frail.

\subsubsection{Study 2B: Interactions are more useful than single-agent traces in expert-scarce settings}

Study 1 showed that learning from pure interaction data is strictly better than learning from pure expert-only data when the task involves recovering from a state that would not have been encountered by an expert in the first place.
    %\item Simulation 2 showed that adding tags enhances performance for trials where the value of interaction is less clear
This raises the question of whether the properties of interaction data make it useful for learning more generic behaviors, namely the task of avoiding high-cost states altogether.
To study this, we constructed a set of with-source \textit{control} datasets, which were matched to with-source interaction datasets (introduced in Study 2A) in the percent of the data that was generated under the expert policy. The key difference was that in the control datasets, the remaining data was generated under the novice policy, rather than the interaction policy (e.g., the control dataset with 5\% expert-only data contained 5000 expert-only traces and 95,000 novice-only traces). 

Using the with-cue method to elicit predictions, we found that training on interaction data yielded higher performance when the proportion of expert-only data was low (5\% or less; Figure \ref{fig:three_panels}B). Beyond that threshold, training on the control data was just as informative. We confirmed that this difference was driven by differences in the content of interaction traces, rather than the amount of training data (see \ref{app:token_control}).
This highlights another, more indirect situation where learning from interaction is beneficial: when interaction surfaces a related, but not directly illustrative behavior (e.g. recovering from a costly state, vs. avoiding costly states altogether). This facilitates learning the target expert behavior (avoiding costly states) when demonstrations of the expert behavior are scarce.
%It may seem surprising that models trained on control datasets can perform well even with modest amounts of expert-only data. However, this can be explained by the observation that in this particular domain, learning the contingencies of novice-only behavior can also be helpful for acquisition of expert-only behavior. Learning to model novices involves learning that only certain kinds of sequences are legal (i.e., agents cannot move diagonally), which can help constrain the learning of expert behavior.

\section{Discussion}
%We examined the factors that predict when learning from traces of interaction confers a benefit beyond learning from a single expert. Our results showed that learning from third-party pedagogical interactions improves performance when exposure to more of the state space is beneficial for the task. Learning from interaction data conferred benefits in additional settings when the data revealed information about the agents generating the observed behavior. %This highlights 2 attributes of interaction data (the presence of corrective actions and source indicators) that could encourage robustness in difficult tasks.
%exposure to interaction data improves performance across a variety of domains: exposure to interactions between experts and novices, with salient indicators of roles of the agents generating the data. 
Here we established a set of key properties that support the effectiveness of interaction data: 1) information asymmetries that create the opportunity for recovery events, which in turn exposes more of the state space in a way that is useful for learning, and 2) a capacity on the part of the learner to represent the distinct agents who generated separate parts of the interaction data. 
An important direction for future work is to validate these findings in more naturalistic settings, e.g. by modifying natural language interaction datasets along the dimensions identified by this analysis as improving performance, and training models on those datasets. These results represent an initial step toward understanding the value of learning from observed interactions, for both humans and AI systems.

%\subsubsection*{Acknowledgments}
%Use unnumbered third level headings for the acknowledgments. All
%acknowledgments, including those to funding agencies, go at the end of the paper.

\bibliography{iclr2026_conference}

@article{chuey2025theory,
  title={Theory of minds: early understanding of interacting minds},
  author={Chuey, Aaron and Gweon, Hyowon},
  journal={Developmental Psychology},
  volume={7},
  number={1},
  pages={91--115},
  year={2025},
  publisher={Annual Reviews}
}

@article{shafto2014rational,
  title={A rational account of pedagogical reasoning: Teaching by, and learning from, examples},
  author={Shafto, Patrick and Goodman, Noah D and Griffiths, Thomas L},
  journal={Cognitive psychology},
  volume={71},
  pages={55--89},
  year={2014},
  publisher={Elsevier}
}

@article{bonawitz2011double,
  title={The double-edged sword of pedagogy: Instruction limits spontaneous exploration and discovery},
  author={Bonawitz, Elizabeth and Shafto, Patrick and Gweon, Hyowon and Goodman, Noah D and Spelke, Elizabeth and Schulz, Laura},
  journal={Cognition},
  volume={120},
  number={3},
  pages={322--330},
  year={2011},
  publisher={Elsevier}
}

@article{foushee2024infants,
  title={Infants who are rarely spoken to nevertheless understand many words},
  author={Foushee, Ruthe and Srinivasan, Mahesh},
  journal={Proceedings of the National Academy of Sciences},
  volume={121},
  number={23},
  pages={e2311425121},
  year={2024},
  publisher={National Academy of Sciences}
}

@article{fox1999listening,
  title={Listening in on monologues and dialogues},
  author={Fox Tree, Jean E},
  journal={Discourse processes},
  volume={27},
  number={1},
  pages={35--53},
  year={1999},
  publisher={Taylor \& Francis}
}

@inproceedings{ross2011reduction,
  title={A reduction of imitation learning and structured prediction to no-regret online learning},
  author={Ross, St{\'e}phane and Gordon, Geoffrey and Bagnell, Drew},
  booktitle={Proceedings of the fourteenth international conference on artificial intelligence and statistics},
  pages={627--635},
  year={2011},
  organization={JMLR Workshop and Conference Proceedings}
}

@inproceedings{du2023improving,
  title={Improving factuality and reasoning in language models through multiagent debate},
  author={Du, Yilun and Li, Shuang and Torralba, Antonio and Tenenbaum, Joshua B and Mordatch, Igor},
  booktitle={Forty-first International Conference on Machine Learning},
  year=2024
}

@article{subramaniam2025multiagent,
  title={Multiagent finetuning: Self improvement with diverse reasoning chains},
  author={Subramaniam, Vighnesh and Du, Yilun and Tenenbaum, Joshua B and Torralba, Antonio and Li, Shuang and Mordatch, Igor},
  journal={arXiv preprint arXiv:2501.05707},
  year={2025}
}

@article{khan2024debating,
  title={Debating with more persuasive llms leads to more truthful answers},
  author={Khan, Akbir and Hughes, John and Valentine, Dan and Ruis, Laura and Sachan, Kshitij and Radhakrishnan, Ansh and Grefenstette, Edward and Bowman, Samuel R and Rockt{\"a}schel, Tim and Perez, Ethan},
  journal={arXiv preprint arXiv:2402.06782},
  year={2024}
}

@article{liu2023training,
  title={Training socially aligned language models on simulated social interactions},
  author={Liu, Ruibo and Yang, Ruixin and Jia, Chenyan and Zhang, Ge and Zhou, Denny and Dai, Andrew M and Yang, Diyi and Vosoughi, Soroush},
  journal={arXiv preprint arXiv:2305.16960},
  year={2023}
}

@article{vafa2024evaluating,
  title={Evaluating the world model implicit in a generative model},
  author={Vafa, Keyon and Chen, Justin Y and Rambachan, Ashesh and Kleinberg, Jon and Mullainathan, Sendhil},
  journal={Advances in Neural Information Processing Systems},
  volume={37},
  pages={26941--26975},
  year={2024}
}

@article{landrum2015learning,
  title={Learning to trust and trusting to learn: A theoretical framework},
  author={Landrum, Asheley R and Eaves, Baxter S and Shafto, Patrick},
  journal={Trends in Cognitive Sciences},
  volume={19},
  number={3},
  pages={109--111},
  year={2015},
  publisher={Elsevier}
}
\bibliographystyle{iclr2026_conference}

\appendix
\setcounter{figure}{0}         % Reset figure counter to 0
\setcounter{table}{0}          % Reset table counter to 0
\renewcommand{\thefigure}{A\arabic{figure}} % Change numbering to A1, A2...
\renewcommand{\thetable}{A\arabic{table}}   % Change numbering to A1, A2...
\section{Appendix}
\subsection{Related work}

\subsection{Formal definition of task}
\label{app:mdp}
Formally, we define a set of Markov Decision Processes (MDPs), where each MDP is defined by the tuple $\langle \mathcal{S}, \mathcal{A}, T, R_{H, g} \rangle$. The state space $\mathcal{S}$ is the set of cells in a 20x20 grid. The action space $\mathcal{A}$ is the set of the cardinal directions $\{N, S, E, W\}$. The transition function $T(s' \vert s, a)$, represents the probability of entering state $s'$ from state $s$ after taking action $a$: here we assume that this function is deterministic, and agents always transition to the intended state. Finally, we define reward function $R_{H, g}(s)$, which is parameterized by the high cost cells $H \subset \mathcal{S}$ and goal state $g$: 
\begin{equation*}
R_{H, g}(s) = 
  \begin{cases} 
  +100 & \text{if } s = g \\
  -20 & \text{if } s \in H \\
  -1 & \text{otherwise}
  \end{cases}
\end{equation*}

\subsection{Grid generation procedure}
\label{app:grid_generation}
We generated 10 unique grids, where the high-cost set $H \subset S$ is constructed by sampling $n \sim \text{Uniform}(1, 4)$ rectangular zones. The dimensions of each zone are sampled as $r, c \sim \text{Uniform}(3, 6)$, and their positions are sampled uniformly at random within the grid bounds.

\subsection{Additional training and evaluation details}
\label{app:training_details}

We used a smaller version of the LLaMa autoregressive transformer architecture, with hidden size layers of length 128, 4 hidden layers and 4 attention heads (a total of approximately 17.6 million parameters). We defined a vocabulary which contained a unique token for each of the 400 unique states as well as a token for each of the 4 possible actions corresponding to the cardinal directions. For all experiments, models were trained on 100,000 generated traces for 10 iterations (to convergence), with a batch size of 16. Separate models were trained for each of the 10 unique grids and 2 policies per grid, for a total of 20 trained models. 
Model predictions were generated using greedy decoding.

\subsection{Additional results}
\subsubsection{Learning from differentiated agents improves
performance under noisy differentiation}
\label{app:noisy_differentiation}
\begin{figure}
    \centering
    \includegraphics[width=0.6\linewidth]{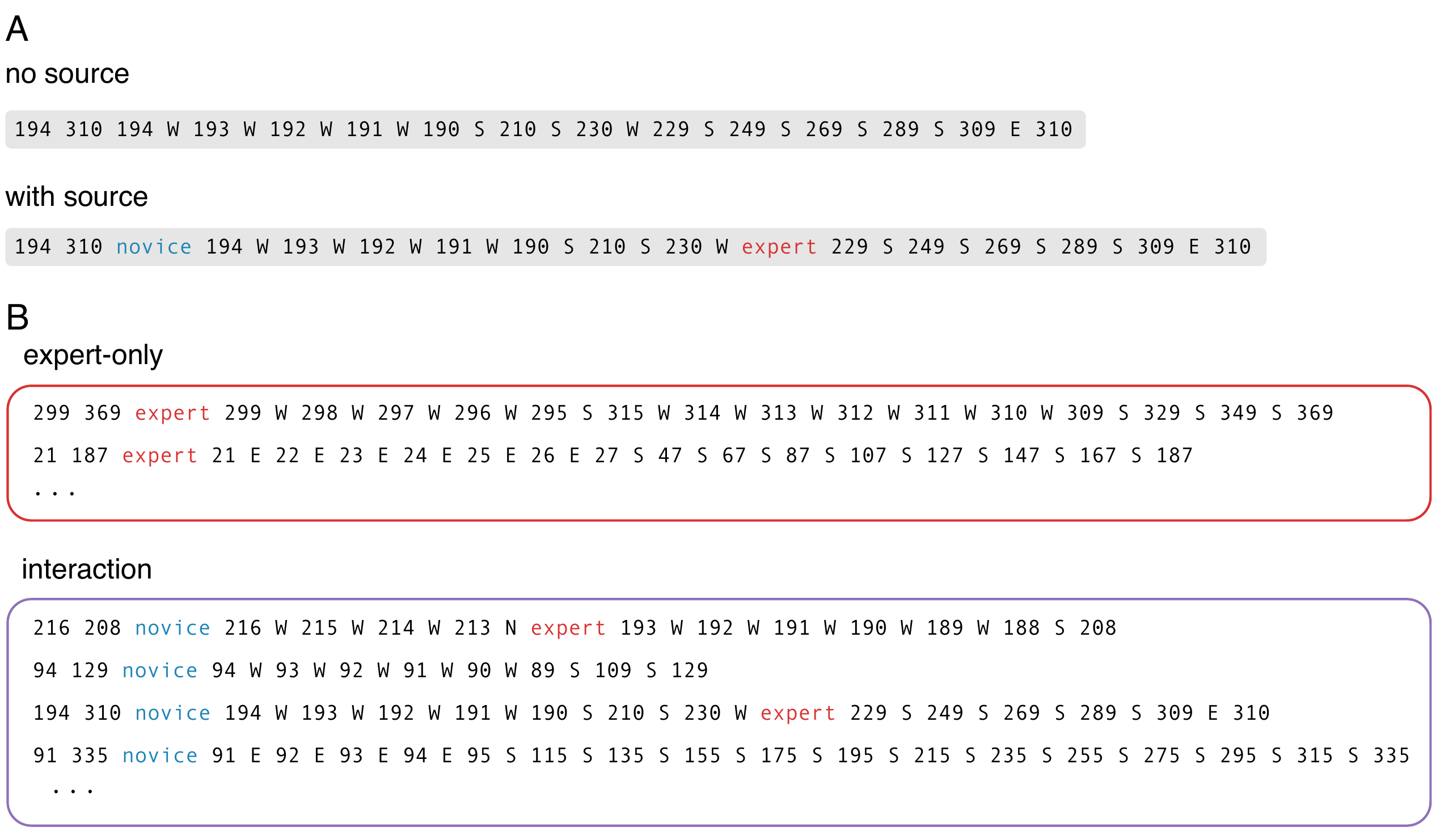}
    \caption{Overview of datasets with source indicators. A: example of a no-source trace and the corresponding with-source trace. B: depiction of an example dataset containing expert-only data and interaction data.}
    \label{fig:tagging_overview}
\end{figure}

The higher performance ceiling of models trained on with-source data highlights the value of information about the type of agent generating the observed data. Yet learning agents may not always have access to unambiguous type information; for example, a true novice may not be able to distinguish between experts and less-skilled individuals. To capture these dynamics, we conducted an additional experiment in which we varied the frequency with which the tags appeared in the training data. Specifically, we reconstructed the training datasets so that for each occurrence of an agent tag, there was a $0 < k < 1$ probability that the tag was visible during training. 

\begin{figure}
    \centering
    \includegraphics[width=0.3\linewidth]{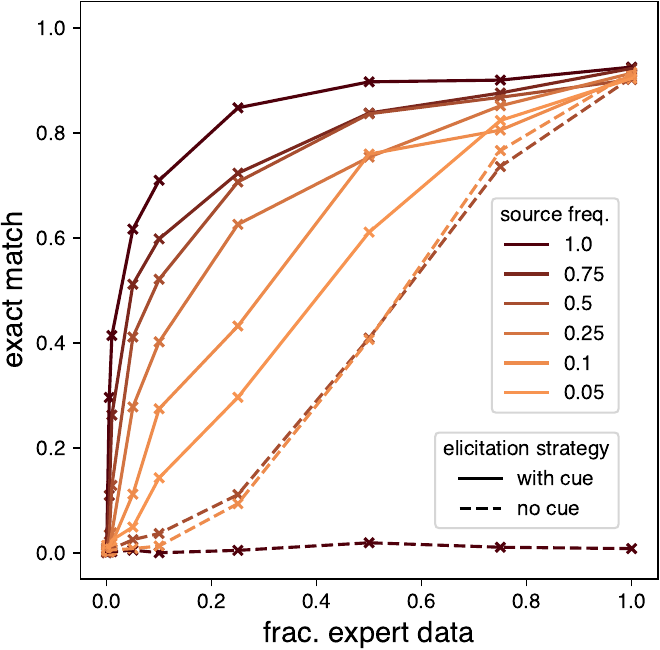}
    \caption{Performance on hazardous trials with varying frequencies of agent indicator tokens, under both prediction elicitation strategies.}
    \label{fig:drop_tags}
\end{figure}

Under the with-cue condition to elicit trajectories, models trained on datasets with partial source information outperformed the linear trend even as the frequency of source indicators was reduced (Figure \ref{fig:drop_tags}). This indicates that in this setting, veridical source information is useful at any volume, rather than being an all-or-nothing phenomenon.
%This highlights that even the sporadic appearance of tags is sufficient to improve model performance when expert-only data is limited.
%When cued without the expert tag, models trained on partially-tagged datasets achieved ceiling performance. This is in contrast to models trained on fully-tagged dataset, which struggle without the expert cue. Overall, this highlights the value of learning from partially-tagged interactions, as it induces 2 modes: using domain knowledge when it is available (e.g. when prompted with expert cue), while defaulting to the most common strategy seen in training in the absence of a cue.

We also investigated how training on datasets with partial rather than full source information affected performance, under the no-cue condition. When $k$, the likelihood of a source indicator token being visible in the data, was less than 1, models reached ceiling performance in the no-cue condition (e.g. performance was linearly proportional to the fraction of the data being generated by an expert). This highlights a key benefit of learning from data with partial, but not full source information: reducing the frequency of source indicators renders learning agents less dependent on these indicators for producing correct trajectories, and therefore more robust in the no-cue condition. These results are consistent with the idea that learning from the partial datasets induces 2 modes: 1) relying on knowledge about the source to generate more accurate predictions when this information is available, in the with-cue condition, and 2) defaulting to the most common strategy seen in training in the absence of a cue.

% Subfigure B

\subsection{Benefits of interaction data over single-agent data result from differences in content, not length}
\label{app:token_control}
Interaction traces contain on average 6\% more tokens than single-agent traces (mean length of a trace: 30.8 vs. 29.1). To confirm that the improvement in performance from training on interaction datasets over single-agent datasets was due to the structural properties of interaction, rather than higher volumes of data, we ran an additional follow-up experiment. We constructed an additional set of interaction datasets that were matched, in the number of expert traces, to the original interaction datasets consisting of 0.5\% expert traces and the control datasets with 0.5\% expert traces (a total of 500 expert traces). The key difference between these new interaction datasets is that the total number of tokens was matched to the total number of tokens in the 0.5\% expert control datasets (approximately $29$ million tokens). Models trained on the token-matched interaction datasets performed substantially better than models trained on the control datasets (Figure \ref{app:token_control}), confirming that the value of interaction data came from its structural properties.

\begin{figure}
    \centering
    \includegraphics[width=0.25\linewidth]{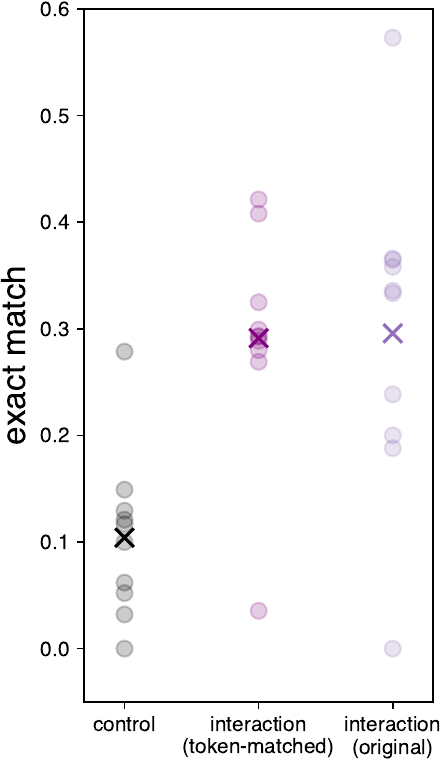}
    \caption{Performance on hazardous trials for models trained on 3 types of datasets: control datasets (500 expert-only traces, 99,500 novice-only traces), token-matched interaction datasets (500 expert-only traces, 93,500 interaction traces; total number of tokens is matched with control datasets), and the original interaction datasets (500 expert-only traces, 99,500 interaction traces).}
    \label{fig:token_control}
\end{figure}
\end{document}